\documentclass{article}
\usepackage{ijcai20}

\usepackage{times}

\usepackage{soul}
\usepackage{url}
\usepackage[hidelinks]{hyperref}
\usepackage[utf8]{inputenc}
\usepackage[small]{caption}
\usepackage{graphicx}
\usepackage{amsmath,amsfonts}
\usepackage{booktabs}
\usepackage[american]{babel}
\usepackage[utf8]{inputenc}

\usepackage{amsmath,amsfonts,bm}

\def\eqref#1{equation~\ref{#1}}

\def\1{\bm{1}}

\DeclareMathAlphabet{\mathsfit}{\encodingdefault}{\sfdefault}{m}{sl}
\SetMathAlphabet{\mathsfit}{bold}{\encodingdefault}{\sfdefault}{bx}{n}

\usepackage{hyperref}
\usepackage{url}
\usepackage{amssymb}
\usepackage{mathtools,algorithm}
\usepackage{enumitem}
\usepackage{hyperref}
\usepackage{subfig}
\usepackage[noend]{algpseudocode}
\makeatletter
\def\BState{\State\hskip-\ALG@thistlm}
\makeatother
\usepackage[capitalise]{cleveref}

\usepackage{todonotes,nth}
\usepackage{multicol}

\definecolor{citecolor}{rgb}{0,0.08,0.45}
\hypersetup{colorlinks,
linkcolor=citecolor,
citecolor=citecolor
}

\newcommand{\removed}[1]{}
\newcommand{\FVE}{\mathcal{V}^{ex}}

\graphicspath{{Gym/}{figures/}{Roboschool/}}

\title{Only Relevant Information Matters:\\Filtering Out Noisy Samples to Boost RL}

\author{
    Yannis Flet-Berliac\footnote{Correspondence to: \href{mailto:yannis.flet-berliac@inria.fr}{\color{black}{yannis.flet-berliac@inria.fr}}} \And Philippe Preux
  \affiliations
  SequeL Inria\\CRIStAL Univ. Lille\\ CNRS\\
}

\begin{document}

\maketitle

\begin{abstract}
  In reinforcement learning, policy gradient algorithms optimize the policy directly and rely on sampling efficiently an environment. Nevertheless, while most sampling procedures are based on direct policy sampling, self-performance measures could be used to improve such sampling prior to each policy update. Following this line of thought, we introduce SAUNA\footnote{\textit{\textbf{S}amples \textbf{A}re \textbf{U}seful? \textbf{N}ot \textbf{A}lways}. Saunas help to release impurities [noisy samples] and improve cell regeneration [PG update]. Their temperatures could be fatal if not regulated by humidity [$\FVE$].}, a method where non-informative transitions are rejected from the gradient update. The level of information is estimated according to the fraction of variance explained by the value function $\FVE$: a measure of the discrepancy between $V$ and the empirical returns. In this work, we use this criterion to select samples that are useful to learn from, and we demonstrate that this selection can significantly improve the performance of policy gradient methods. In this paper: (a) We define $\FVE$ and introduce the SAUNA method to filter transitions. (b) We conduct experiments on a set of benchmark continuous control problems. SAUNA significantly improves performance. (c) We investigate how $\FVE$ reliably selects samples with the most positive impact on learning and study its improvement on both performance and sample efficiency.
\end{abstract}

\section{Introduction}
Learning to control agents in simulated environments has been a challenge for decades in reinforcement learning (RL)~\cite{werbos1989neural,Robinson1989DynamicRD,schmidhuber1991learning} and has lately led to a lot of research efforts in this direction~\cite{silver_mastering_2016,ha2018recurrent,espeholt2018impala}, notably in policy gradient methods~\cite{silver2014deterministic,schulman2015high,haarnoja2018soft}. Despite progress, policy gradient algorithms still heavily suffer from sample inefficiency~\cite{kakade2003sample,wang2016sample,wu2017scalable}. In particular, many of those methods are subject to use as much experience as possible in the most efficient way. However, quantity is not quality: the quality of the sampling procedure also determines the learning curve of the agent and its final performance. Hence, we think that \textit{not all experiences are worth using} in the gradient update. Indeed, some transitions may add noise to the gradient update, diluting relevant signals, and hindering learning. The central idea of SAUNA is to reject transitions that are not informative.

Use of non-informative or misinformative transitions can only mislead the learning process and waste computational time. Amari's natural gradient~\cite{amari:natural} concept concerns the geometry of the search space related to the ``value of information'': this has been studied for long in RL since~\cite{kakade2002natural}. Our work focuses on a different notion of value of information, and treats it differently: we evaluate whether a transition conveys useful information and use it only if it is considered beneficial to learning. For this purpose, we use a measure of the discrepancy between the estimated state value and the observed returns. This discrepancy is formalized with the notion of the fraction of variance explained $\FVE$~\cite{10.2307/2683704}. Transitions for which $\FVE$ is close to zero are those for which the correlation between the value function $V$ and the observed returns is also close to zero. SAUNA keeps transitions where there is either a strong correlation or a lack of fit between $V$ and the returns while avoiding the dilution of useful information by removing useless samples. We consider on-policy methods for their unbiasedness and stability compared to off-policy algorithms~\cite{nachum2017bridging}. However, our method can be applied to off-policy methods as well, and we leave this investigation open for future work.

In summary, in this paper:
\begin{enumerate}
\item We propose to move from a traditional policy-based sampling procedure to a refined sample selection driven by $\FVE$. We explore how transition filtering simplifies the underlying state space and affects performance.

\item We hypothesize that not all samples are useful for learning and that disturbing samples should be rejected to avoid performance loss. We provide experimental evidence corroborating this claim.

\item By combining (1) and (2), we obtain a learning algorithm that is empirically effective in learning neural network policies for challenging control tasks. Our results significantly improve the state of the art in using RL for high-dimensional continuous control.
\end{enumerate}
Section \ref{sec:preliminaries} recalls basic notions of policy gradient methods in RL and the notion of ``fraction of variance explained'' drawn from the statistics literature. Section \ref{sec:relatedWork} sets our contribution within the RL domain. Section \ref{sec:method} introduces SAUNA. Section \ref{sec:exp} provides experimental evidence of the benefit of using our method, and also investigates various experimental aspects of SAUNA. Section \ref{sec:disc} further discusses the method. Finally, Section \ref{sec:conclusion} concludes and draws some lines of future research.

\section{Preliminaries}
\label{sec:preliminaries}
\subsection{Notations}
We consider a Markov Decision Process (MDP) with states $s \in \mathcal{S}$, actions $a \in \mathcal{A}$, transition distribution $s_{t+1} \sim {\cal P} (s_{t},a_{t})$ and reward function $r_t \sim {\cal R} (s_t, a_t)$. Let $\pi(a | s)$ denote a stochastic policy and let the objective function be the expected sum of discounted rewards:
\begin{equation}
J(\pi) \triangleq \underset{\tau \sim \pi}{\mathbb{E}}\left[\sum_{t=0}^{\infty} \gamma^{t} r\left(s_{t}, a_{t}\right)\right],
\end{equation}
where $\gamma \in [0,1)$ is a discount factor~\cite{puterman1994markov} and $\tau=\left(s_0, a_0, r_0, s_1, a_1, r_1, \dots\right)$ is a trajectory sampled from the environment while the agent is following a given policy $\pi$. Let us remind the notions of the value of a state in the MDP framework. The value $V^\pi(s)$ of a state $s$ while following a policy $\pi$ starting in state $s$ is defined by: $V^\pi(s) \triangleq \underset{\tau \sim \pi}{\mathbb{E}}\left[\sum_{t=0}^{\infty} \gamma^{t} r\left(s_{t}, a_{t}\right) | s_0 = s\right]$.

Closely related is the value (or quality) of a state-action pair: the quality $Q^\pi (s,a)$ of performing action $a$ in state $s$ and then following policy $\pi$ is defined by: $Q^\pi (s,a) \triangleq \underset{\tau \sim \pi}{\mathbb{E}}\left[\sum_{t=0}^{\infty} \gamma^{t} r\left(s_{t}, a_{t}\right) | s_0 = s, a_0 = a\right]$. Finally, the advantage function quantifies how an action $a$ is better than the average action in state $s$ (following policy $\pi$): $A^\pi (s,a) \triangleq Q^\pi (s,a) - V^\pi (s)$. MDP theory asserts that there exists an optimal policy $\pi^*$ that maximizes $J$: we denote its value function $V^*$. In practice, value functions are unknown; we denote $V$, $Q$, and $A$ their current estimates.

\subsection{Policy Gradient Methods}
Policy gradient methods aim at optimizing the policy directly~\cite{williams1992simple}.
The policy $\pi$ is often implemented with a function parameterized by $\theta$: learning a policy boils down to finding the best parameters. In the sequel, we use $\theta$ to denote the parameters as well as the policy. In deep RL, the policy is represented by a neural network (the policy network) and is assumed to be continuously differentiable with respect to its parameters $\theta$. When the policy is represented by such a parameterized function, hence by an approximation of a policy, the MDP theory basically breaks down.

In this paper, we consider Proximal Policy Optimization (PPO)~\cite{schulman2017proximal}, an on-policy policy gradient method achieving state of the art performance on a suite of benchmark tasks despite a relatively simple implementation. Very interestingly, PPO is an evolution of TRPO that builds on the notion of natural gradient, hence Amari's notion of ``value of information'' mentioned above. PPO has been shown to outperform TRPO experimentally. By building on PPO, this paper combines two different ideas related to the notion of the value of information. At each episode, PPO collects $(s_t, a_t, r_t)$ samples using its current policy $\theta_k$. After some episodes, using these collected transitions, PPO updates its policy and gets a new one $\theta_{k+1}$:

\begin{equation}
  \theta_{k+1} \gets \underset{\theta}{\mathrm{argmax}} \underset{s_{t}, a_{t} \sim \pi_{\theta_k}}{\mathbb{E}}\left[\mathcal{L}_\text{PPO}\left(s_{t}, a_{t}, \theta_k, \theta\right)\right].
\end{equation}
We use the clipped version of PPO:
\begin{equation}
  \mathcal{L}_\text{PPO}(s_t,a_t,\theta_k,\theta) = 
   \mbox{Clip} (A^{\pi_{\theta_k}}(s_t,a_t),
                 \frac{\pi_{\theta}(a_t|s_t)}{\pi_{\theta_k}(a_t|s_t)},
                 \delta),
  \label{eq:min}
\end{equation}
where $
 \mbox{Clip} (A, \alpha, \delta) = \left\{
  \begin{array}{ll}
    \min{(\alpha A, (1+\delta) A)}, A \geq 0 \\
    \min{(\alpha A, (1-\delta) A)}, A < 0.
  \end{array}\right.
$
$A$ is the advantage function introduced above. Clipping makes the training updates more stable: it ensures that the gradient steps do not lead the policy outside of the region of parameter space where the samples collected are informative.

\subsection{$\FVE$: Fraction of Variance Explained}
Now we introduce the key notion of this paper, namely the fraction of variance explained, denoted by $\FVE$. As shown in this paper, a yet elementary use of this concept strikingly improves the performance of policy gradient algorithms. In general terms, $\FVE$ gives some information about the goodness of fit of a model. In statistics, it is also denoted $R^2$, which is a poor notation since this quantity can be negative for non-linear models~\cite{10.2307/2683704} (also, in the context of RL, $R$ usually refers to the return). This quantity is also known as the coefficient of determination. In a regression setting, assume a model $\hat{y}$ aims at predicting $y$ from $x$, given a set of $N$ couples $(x_i, y_i)$, $\FVE$ is defined by:
  \begin{equation}
    \FVE \triangleq 1 - \frac{\mathrm{MSE}}{\mathrm{VAR}}
  \end{equation}
where $\mathrm{MSE}$ is the mean squared error of the model measured on these $N$ couples ($\mathrm{MSE} = \frac{1}{N}\sum_i \left(y_i - \hat{y} (x_i)\right)^2$), and $\mathrm{VAR}$ is the variance of the observed targets $y_i$. $\FVE \le 1$ and:
\begin{itemize}
\item $\FVE = 1$ means that the model perfectly predicts the data ($\mathrm{MSE} = 0$).
\item $\FVE = 0$ means that the model performs as always predicting the average ($\mathrm{MSE} = \mathrm{VAR}$).
\item $\FVE < 0$ means that the model performs worse than merely predicting the mean value ($\mathrm{MSE} > \mathrm{VAR}$).
\end{itemize}

\section{Related Work}
\label{sec:relatedWork}
Our method integrates three key ideas: (a) function approximation with a neural network combining or separating the actor and the critic with an on-policy setting, (b) transition filtering reducing information/signal dilution in the gradient update while simplifying the underlying MDP, and (c) using $\FVE$ as a measure of correlation between the value function and the returns to allow better sampling and more efficient learning. Below, we consider previous work building on some of these approaches.

Actor-critic algorithms essentially use the value function to alternate between policy evaluation and policy improvement~\cite{barto1983neuronlike,sutton2018reinforcement}. In order to update the actor, many methods adopt the on-policy formulation~\cite{peters2008reinforcement,mnih2016asynchronous,schulman2017proximal}. However, despite their important successes, these methods suffer from sample complexity.

In the literature, research has also been conducted in prioritization sampling. While \cite{schaul2015prioritized} makes the learning from experience replay more efficient by using the TD error as a measure of these priorities in an off-policy setting, our method directly selects the samples on-policy. \cite{schmidhuber1991curious} is related to our method in that it calculates the expected improvement in prediction error, but with the objective to maximize the intrinsic reward through artificial curiosity. Instead, our method estimates the expected fraction of variance explained and filters out some of the samples to improve the learning efficiency.

$\FVE$ has already been used in~\cite{merl} as one of the auxiliary tasks for self-assessment of performance. Finally, motion control in physics-based environments is a long-standing and active research field. In particular, there are many prior work on continuous action spaces~\cite{levine2014learning,heess2015learning,lillicrap2015continuous,schulman2015high} that demonstrate how locomotion behavior and other skilled movements can emerge as the outcome of optimization problems.

\section{SAUNA: Dynamic Transition Filtering}
\label{sec:method}
We introduce a general method to filter transitions that contain useful information for policy gradient updates. In this paper, we detail how to couple SAUNA with PPO, an on-policy gradient algorithm achieving state of the art performance. We refer to this combination as PPO+SAUNA. SAUNA can be coupled with other algorithms, especially with non-policy methods such as DQN: we leave this for future work. Below we detail how to adapt the notion of $\FVE$ to RL.

\subsection{$\FVE$ applied to RL}
The fraction of variance that the current estimate of the value function explains about the observed returns corresponds to the proportion of the variance in the dependent variable $V$ that is predictable from $s_{t}$. We define $\FVE_{\tau}$ as the fraction of variance explained for a trajectory $\tau$:

\begin{equation}
  \FVE_{\tau} \triangleq 1 - \frac{\sum_{t\in\tau}\left(R_t-V(s_{t})\right)^{2}}{\sum_{t\in\tau}\left(R_t- \langle R \rangle_\tau \right)^{2}},
  \label{eq:vex}
\end{equation}
where $R_t = \sum_{k\ge{}0} \gamma^k r_{t+k}$, $r_t$ is the immediate reward collected at timestep $t$, $V(s_{t})$ is the current estimate of the value of state $s_t$, and $\langle R \rangle_\tau$ is the average of the $R_t$ in trajectory $\tau$. This definition can be extended from a trajectory $\tau$ to a batch ${\cal B}$ of sampled transitions $\FVE_{{\cal B}}$. In the RL context, the interpretation of $\FVE_{{\cal B}}$ is:

\begin{itemize}
\item $\FVE_{{\cal B}} = 1$: $V$ perfectly explains the observed returns.
\item $\FVE_{{\cal B}} = 0$: $V$ corresponds to a simple average prediction.
\item $\FVE_{{\cal B}} < 0$: $V$ provides a worse prediction than the average of the returns.
\end{itemize}

The intuition is that $\FVE_{{\cal B}}$ close to 1 corresponds to well-predicted returns. $\FVE_{{\cal B}} < 0$ corresponds to a rather large prediction error of the value function, meaning that these samples are useful because the agent has something to learn from. On the other hand, $\FVE_{{\cal B}}$ close to 0 means that the samples do not provide any valuable information to improve the value estimates. We will demonstrate that $\FVE$ is indeed a relevant indicator for assessing self-performance in RL.

\subsection{Estimating $\FVE$}
While sampling the environment, SAUNA rejects transitions for which $V(s_{t})$ is not correlated with returns that have followed $s_{t}$. Therefore, $\FVE_{{\cal B}}$ should be estimated at each timestep and we define $\FVE_{\theta}(s_{t})$ as the prediction of $\FVE_{{\cal B}}$ with parameters $\theta$ at state $s_{t} \in {\cal B}$. In addition, for shared parameters configurations, an error term on the value estimation is added to the objective. The final objective function becomes:
\begin{align}
  \mathcal{L}_{\mathrm{SAUNA}}(s_{t},a_{t},\theta_{old},\theta)=&\;\mathcal{L}_\text{PPO}(s_{t},a_{t},\theta_{old},\theta)-\label{eq:PPO1}\\
    &\;c_{1}\left(V_{\theta}(s_{t})-R_{t}\right)^{2}-\label{eq:PPO2}\\
    &\;c_{2}\left(\FVE_{\theta}(s_{t})-\FVE_{{\cal B}}\right)^{2}
    \label{eq:objVex}, 
\end{align}
where $c_{1}$ and $c_{2}$ are the coefficients for the squared-error losses of respectively the value function and the fraction of variance explained function. Note that only the term (\ref{eq:objVex}) is specific to SAUNA. (\ref{eq:PPO1}) and (\ref{eq:PPO2}) come from PPO. When the network is not shared between the policy and the value function, SAUNA embeds $\FVE_{{\cal B}}$ to the value function network using a single hidden layer. The rest of the network is unchanged, making our method very easy to use without significantly increasing the complexity of the underlying algorithm.

\subsection{SAUNA Algorithm}
\cref{alg:learning} shows the pseudocode of SAUNA when coupled with PPO. Overall, the resulting algorithm visits a set of trajectories along which it collects useful samples in the sense explained above, assessed with regards to $\FVE$. The mechanism may be viewed as analogous to the method of dropout in deep learning~\cite{srivastava2014dropout,learningtopredict2019} although here dropout happens in the state space of the underlying MDP and is directed by $\FVE$. Once a batch ${\cal B}$ of $T$ such useful samples is collected, SAUNA performs the usual gradient update following the PPO template.

The gradient update concerns the three quantities estimated by SAUNA: the policy parameters $\theta$, line \ref{eq:updatethetaInAlgo}, the value estimation parameters $\phi$, line \ref{eq:updatephiInAlgo}, and the $\FVE$ estimation parameters $\psi$, line \ref{eq:updatepsiInAlgo}. The \textit{if} statement filters the useful samples: $\widetilde{\FVE_{\psi_{k}}}(s_{0:t-1})$ denotes the median of $\FVE_{\psi_{k}}$ between timesteps $0$ and $t-1$, $\epsilon_{0}$ is a Laplace estimator (set to $10^{-8}$), and $\rho$ is the filtering threshold. One may legitimately ask why not use directly $|\FVE_{\psi_{k}}(s_{t})|$ in the predicate. The rationale is practical: the ratio is a standardized measure as the agent learns, stabilized by the median, more robust to outliers than the mean. For better legibility, \cref{alg:learning} does not share parameters between the $\pi$, $V$ and $\FVE$ networks. A version where these parameters would be partially shared is straightforward.

\def\HS{\hspace{\fontdimen2\font}}
\begin{algorithm}
  \caption{SAUNA coupled with PPO.}\label{alg:learning}
\begin{algorithmic}[1]
\State \textbf{Initialize} policy parameters $\theta_{0}$, value function parameters $\phi_{0}$ and $\FVE$ function parameters $\psi_{0}$
\For {$k = 0,1,2,\ldots$}
\State $s_0 \gets$ initial state 
\State batch ${\cal B} \gets \emptyset$
\While {size$({\cal B})\leq{}T$}
\State $a_{t} \sim \pi_{\theta_{k}}(s_{t})$
\State \textbf{execute} action $a_{t}$ and observe $r_{t+1}$ and $s_{t+1}$
\If {$\frac{|\FVE_{\psi_{k}}(s_{t})|}{|\widetilde{\FVE_{\psi_{k}}}(s_{0:t-1})|+\epsilon_{0}} \geq \rho$}\label{line:if}
\State \textbf{add} $\left(s_{t}, a_{t}, r_{t}, V_{\phi_{k}}(s_{t}), s_{t+1}, \FVE_{\psi_{k}}(s_{t})\right)$ to ${\cal B}$
\EndIf
\If {$s_{t+1}$ is a final state}
\State $s_{t+1} \gets$ initial state
\EndIf
\EndWhile
\State $\theta_{k+1} \gets\HS\HS\HS \underset{\theta}{\mathrm{argmax}} \sum_{t\in{\cal B}} \mathcal{L}_\text{PPO}\left(s_{t}, a_{t},\theta_k, \theta\right)$\label{eq:updatethetaInAlgo}
\State $\hphantom{\theta_{k+1}}\mathllap{\phi_{k+1}} \gets\HS\HS\HS \underset{\phi}{\mathrm{argmin}} \sum_{t\in{\cal B}}\left(V_{\phi}\left(s_{t}\right)-R_{t}\right)^{2}$\label{eq:updatephiInAlgo}
\State $\hphantom{\theta_{k+1}}\mathllap{\psi_{k+1}} \gets\HS\HS\HS \underset{\psi}{\mathrm{argmin}} \sum_{t\in{\cal B}}\left(\FVE_{\psi}\left(s_{t}\right)-\FVE_{{\cal B}}\right)^{2}$\label{eq:updatepsiInAlgo}
\EndFor
\end{algorithmic}
\end{algorithm}

\section{Experiments}
\label{sec:exp}
We have forked the \textit{stable-baselines} repository~\cite{stable-baselines} and minimally modified the code to incorporate our method. Unless otherwise stated, the policy network used for all tasks is a fully-connected multi-layer perceptron with 2 hidden layers of 64 units. Moreover, the architecture for the $\mathcal{V}^{ex}$ function head is the same as for the value function head.

\subsection{SAUNA in the Continuous Domain}
To assess SAUNA, we compare PPO+SAUNA against its natural baseline PPO. We use six simulated robotic deterministic tasks from OpenAI Gym~\cite{brockman2016openai} using~\textit{MuJoCo}~\cite{todorov2012mujoco}.
\begin{figure}[h]
    \centering
    \subfloat{{\includegraphics[width=.47\linewidth]{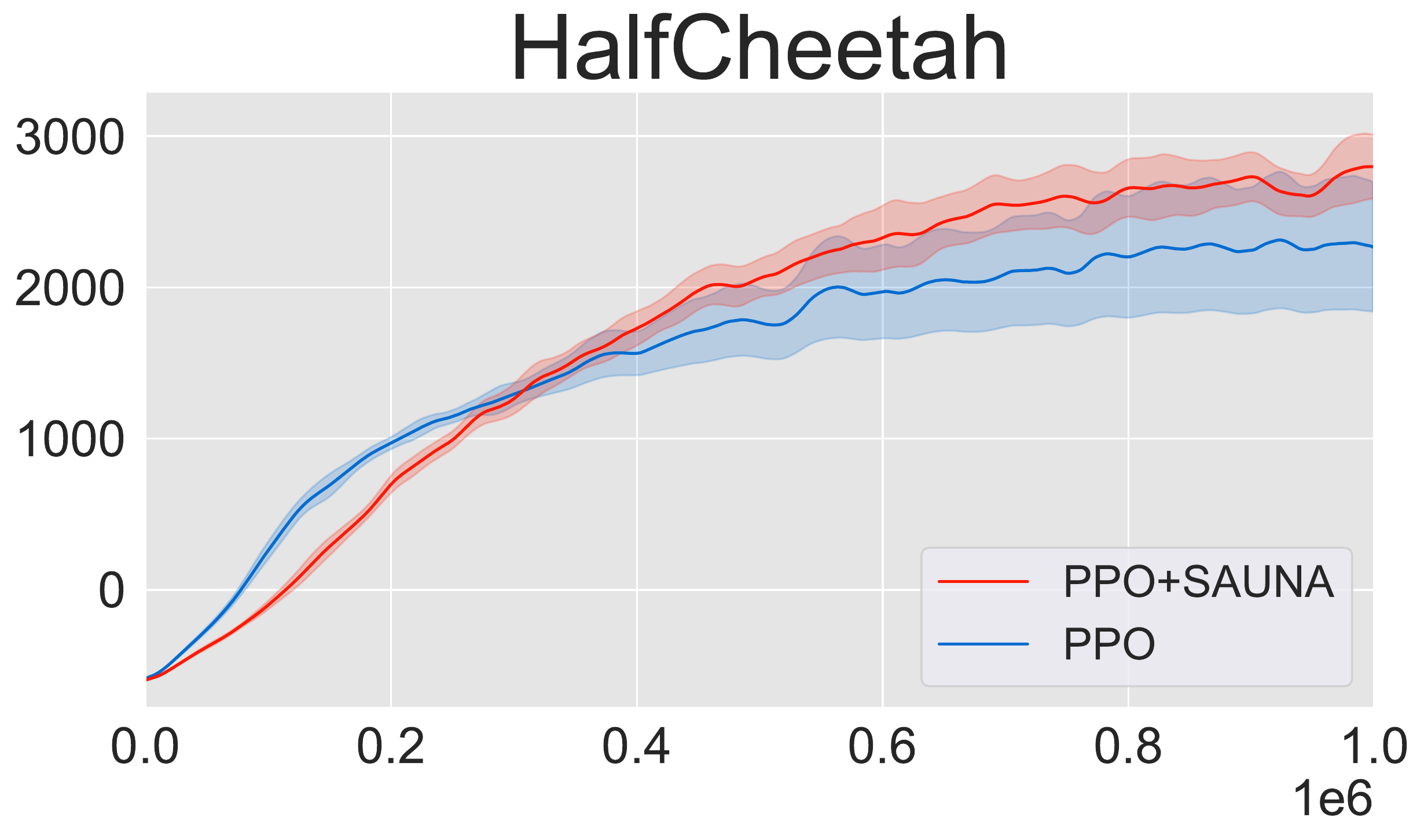}}{\includegraphics[width=.47\linewidth]{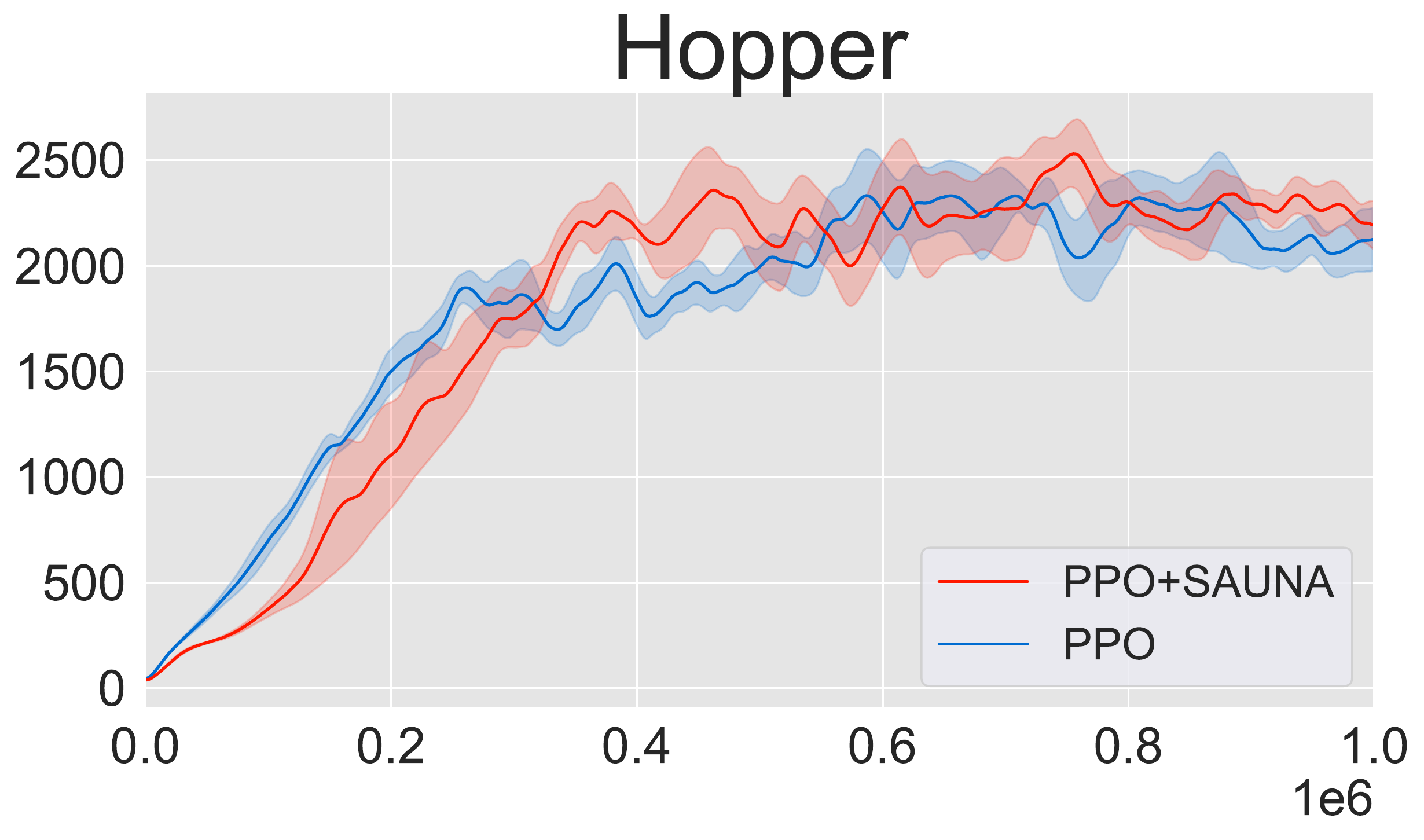}}}
    \qquad
    \subfloat{{\includegraphics[width=.47\linewidth]{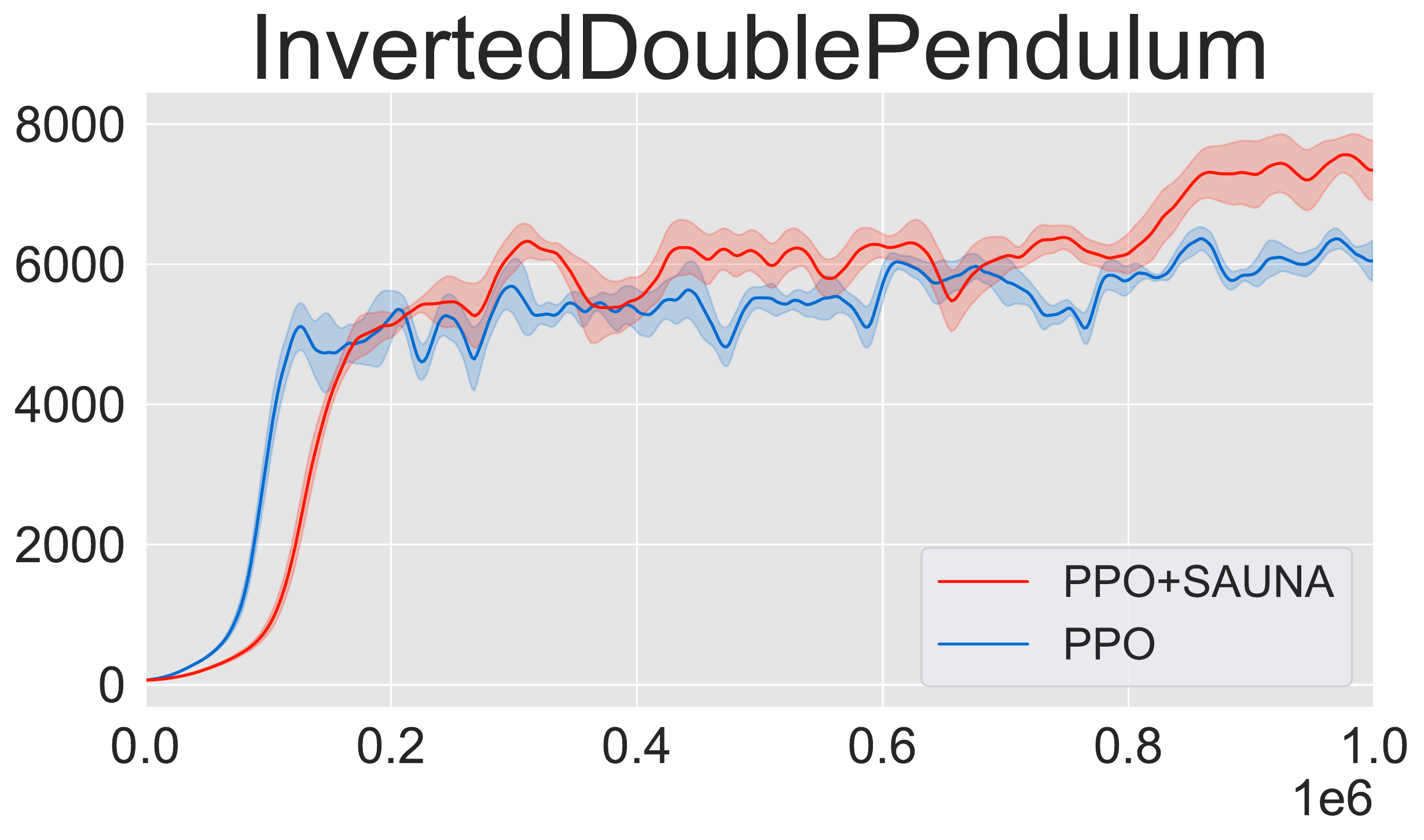}}{\includegraphics[width=.47\linewidth]{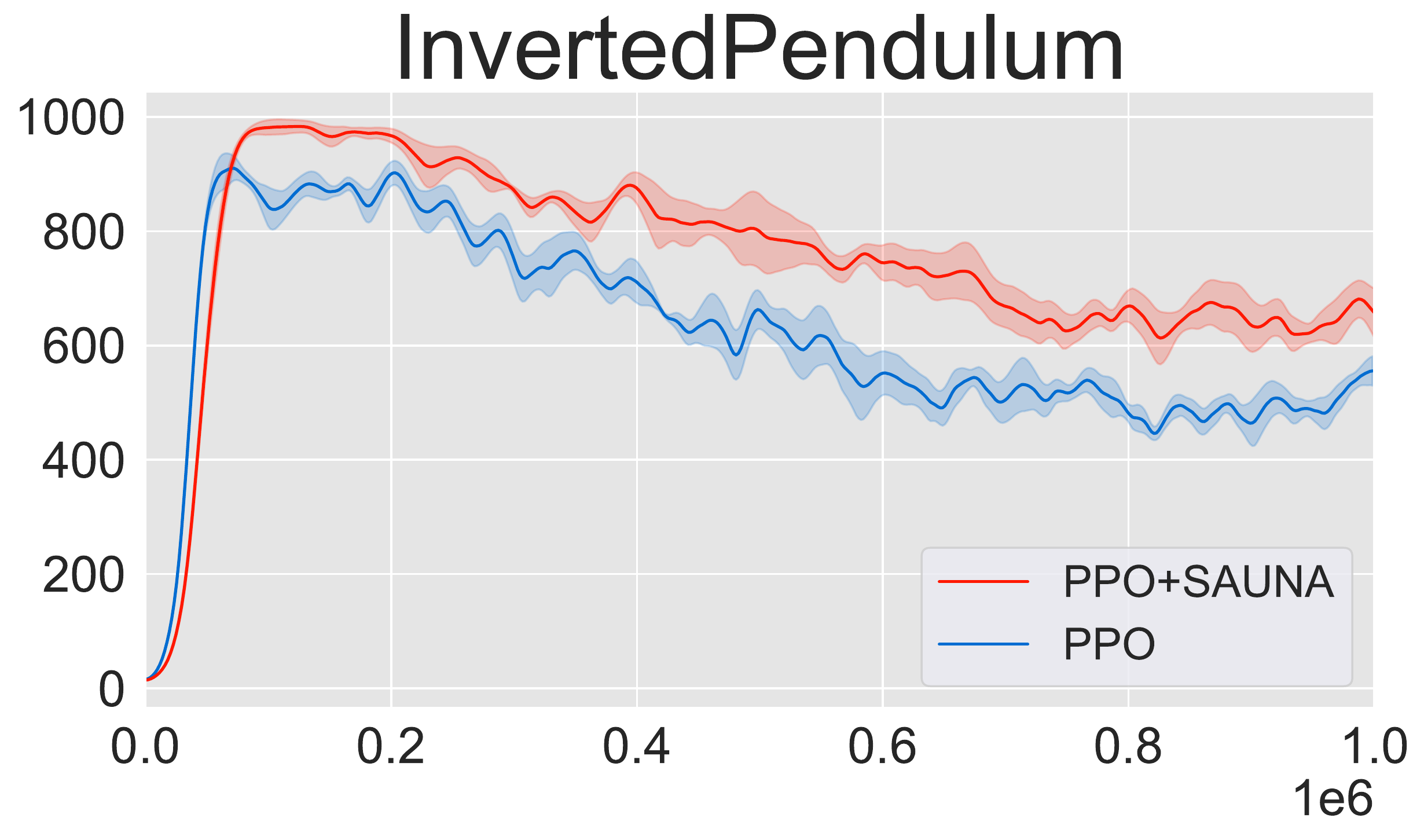}}}
    \qquad
    \subfloat{{\includegraphics[width=.47\linewidth]{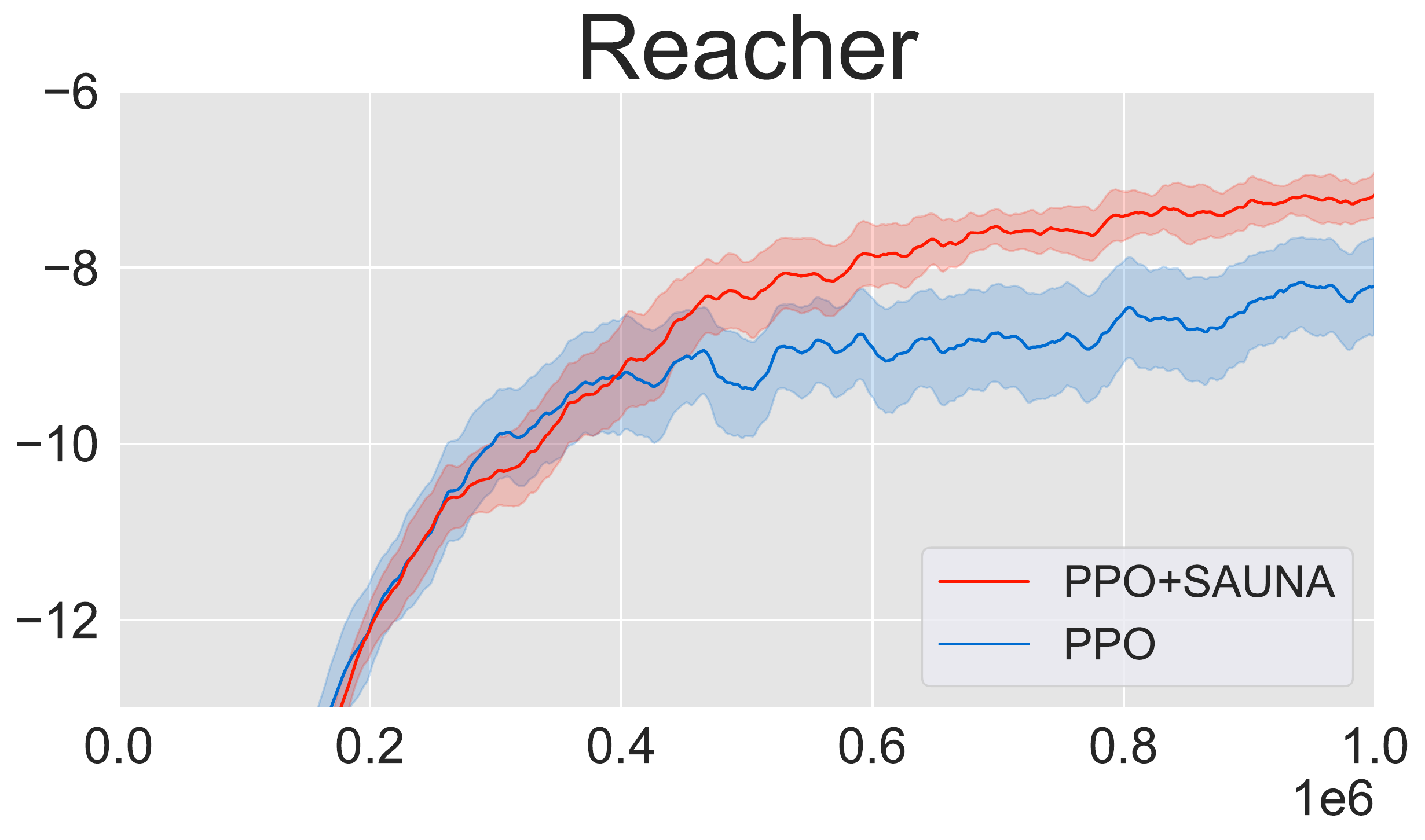}}{\includegraphics[width=.47\linewidth]{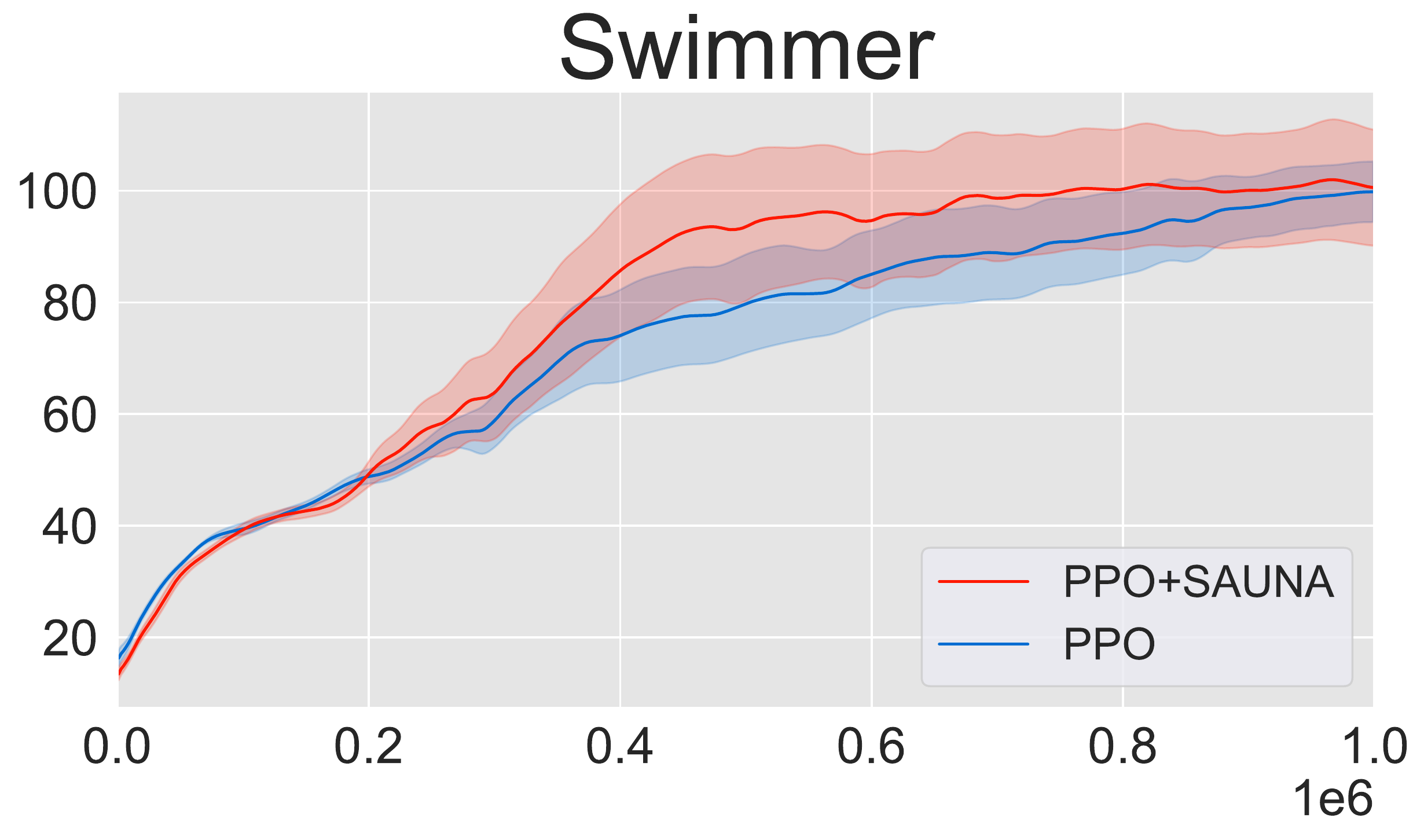}}}
    \caption{Performance of PPO+SAUNA (red) relative to PPO (blue) on 6 \textit{MuJoCo} environments averaged across 6 seeds. X-axis: number of environment steps. Y-axis: total undiscounted return. Shaded areas: standard deviation.}
    \label{fig:mujoco}
\end{figure}The two hyperparameters required by our method ($\rho=0.3$ from Eq.~\ref{eq:vex} and $c_{2}=0.5$ from Eq.~\ref{eq:objVex}) and all the others (identical to those in~\cite{schulman2017proximal}) are exactly the same for all tasks.

We made the choice of not tuning the hyperparameters for each algorithm and for each task to have a tougher assessment of SAUNA: only SAUNA-specific hyperparameters $\rho$ and $c_2$ have been tuned by grid-search. Hence, the performance we report for SAUNA is not necessarily the best that could be obtained with parameter tuning. The graphs reported in~\cref{fig:mujoco} show that our method outperforms PPO on all considered continuous control tasks.

\begin{figure}[h]
  \centering
  \subfloat[]{\includegraphics[width=.47\linewidth]{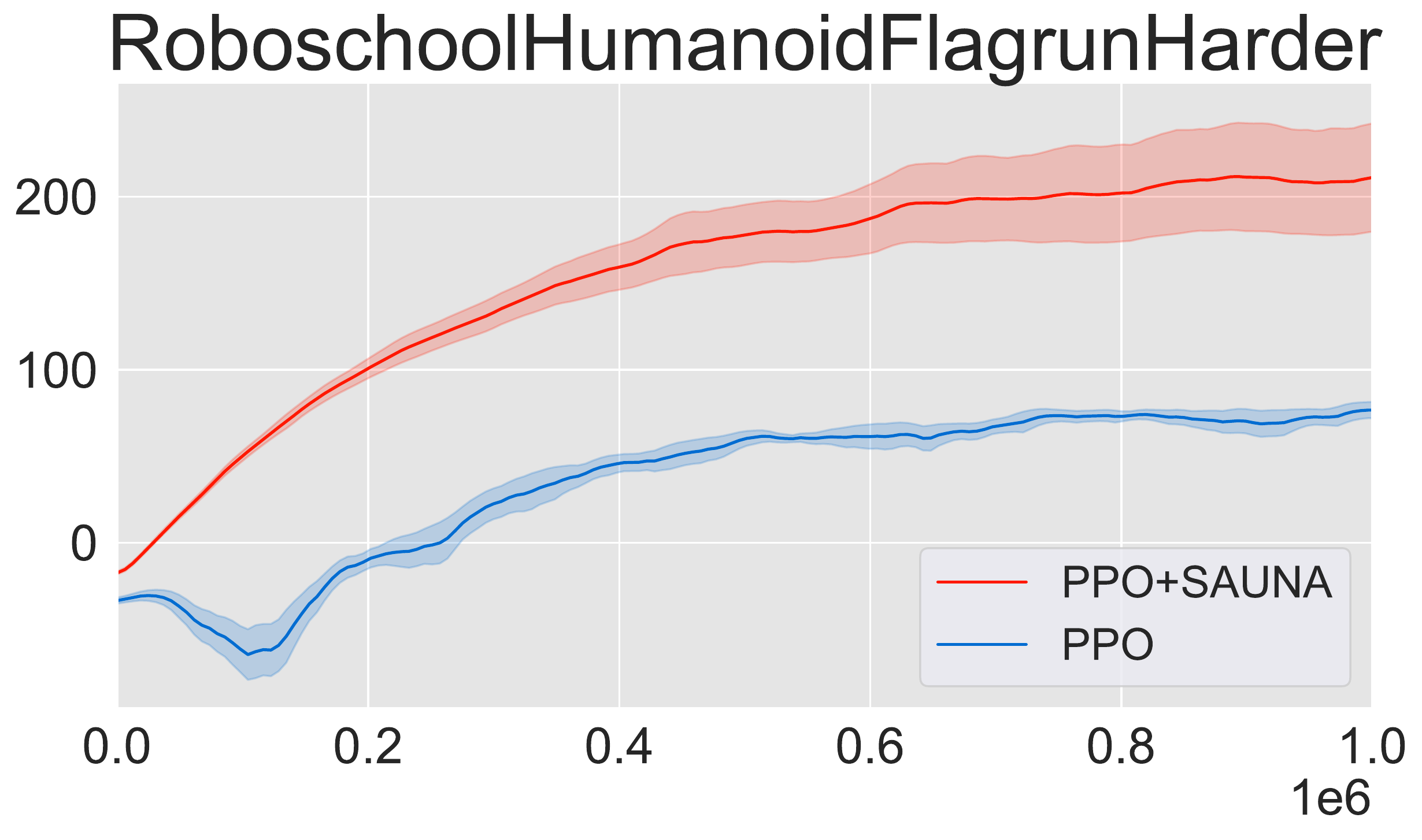}\label{fig:roboschool-a}}\hspace{-0.7cm}
  \qquad
  \subfloat[]{\includegraphics[width=.47\linewidth]{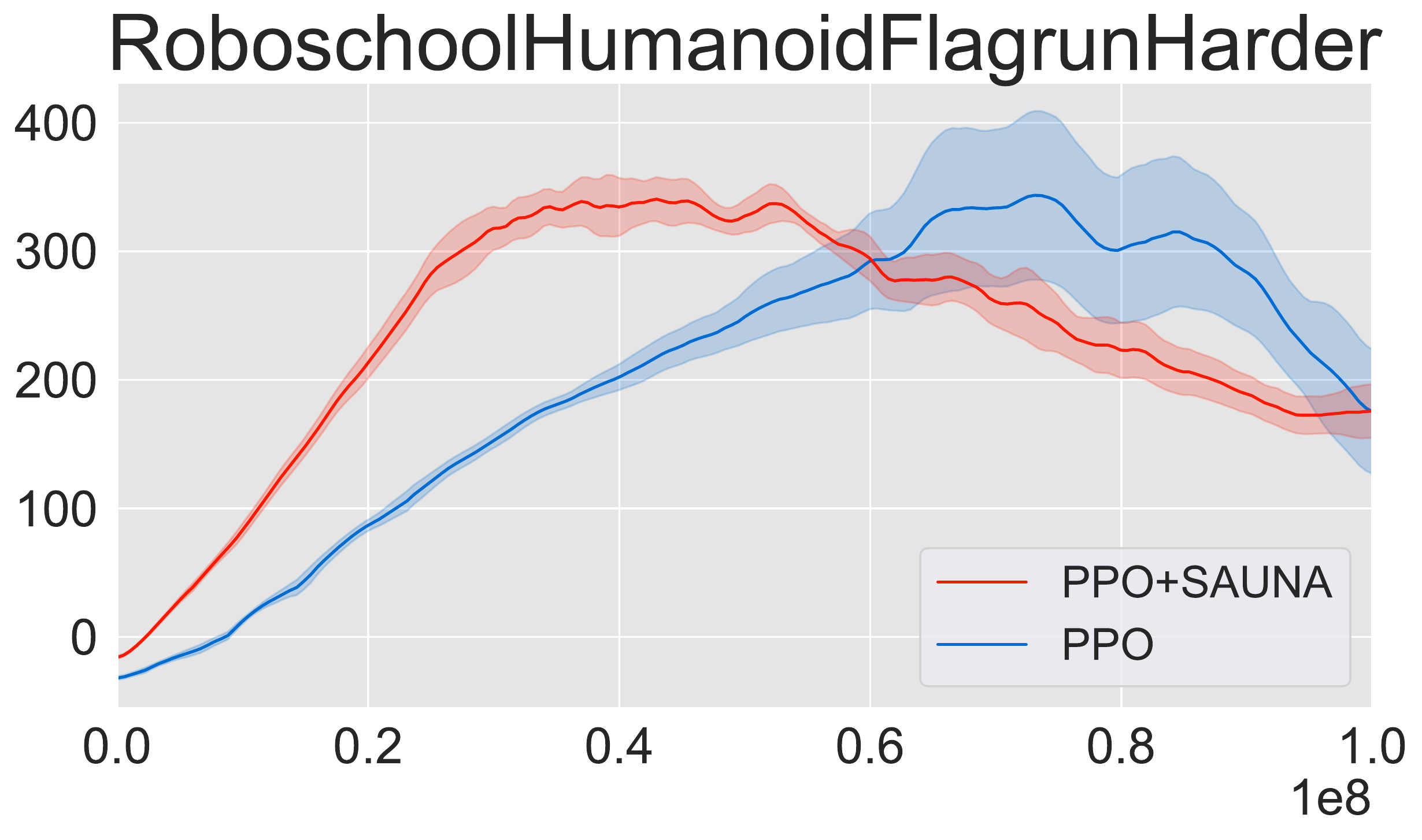}\label{fig:roboschool-b}}\\[-1ex]
  \caption{Performance of PPO+SAUNA (red) relative to PPO (blue) on the \textit{Roboschool} environment averaged across 6 seeds. X-axis: number of environment steps. Y-axis: total undiscounted return. Shaded areas: standard deviation.}
  \label{fig:roboschool}
\end{figure}

We then experiment with the more difficult, high-dimensional continuous domain environment of \textit{Roboschool}~\cite{klimov2017roboschool} with various neural network sizes. In~\cref{fig:roboschool-a}, the same fully-connected network as for the previous \textit{MuJoCo} experiments (2 hidden layers each with 64 neurons) is used. In~\cref{fig:roboschool-b}, the network is composed of a deeper and wider 3 hidden layers with 512, 256 and 128 neurons. We trained those agents with 32 parallel actors. In both experiments, PPO+SAUNA performs better and learns faster at the beginning. The gap closes with a larger network and our method does as well as PPO. As resources are limited in terms of the number of parameters and models become less complex, it seems natural that filtering samples according to their expected informational value helps to reduce noise in the gradient update and to speed up learning.


\subsection{Learning with SAUNA}
\subsubsection{The Advantages of Filtering}
\label{sec:exp_filter}
\begin{figure*}[h]
  \centering
  \subfloat{{\includegraphics[width=.33\linewidth]{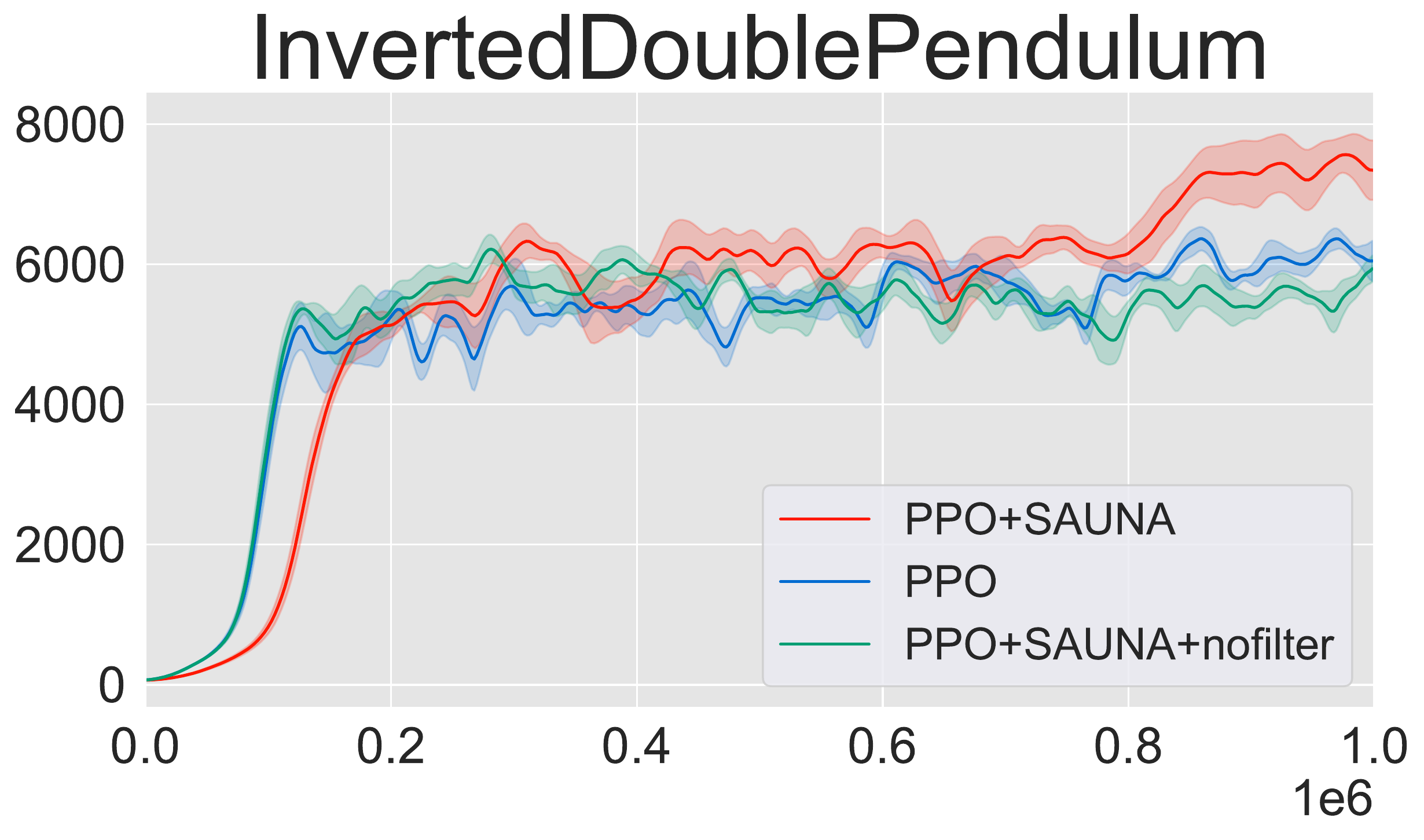}}{\includegraphics[width=.33\linewidth]{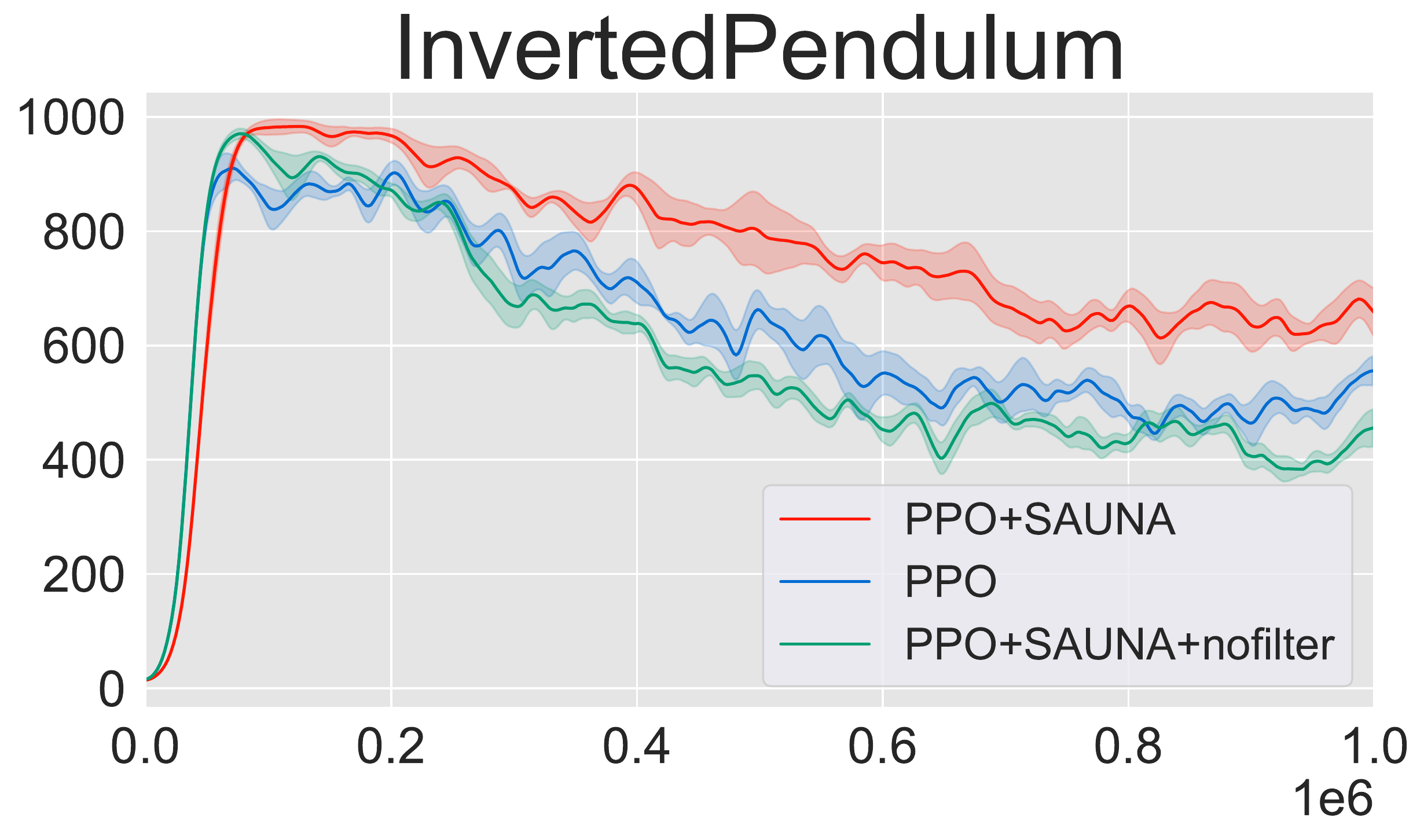}}{\includegraphics[width=.33\linewidth]{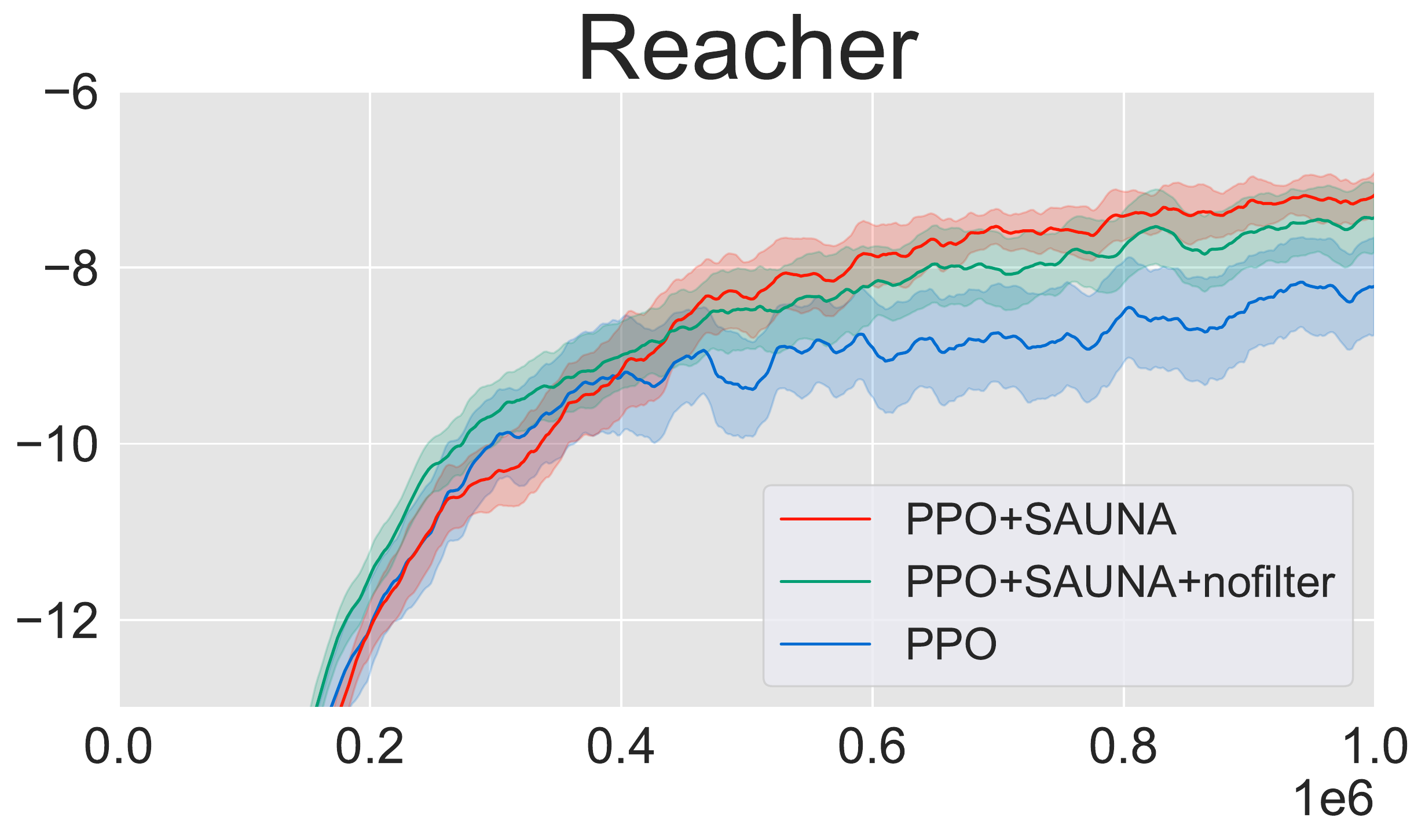}}}
  \caption{Performance of PPO+SAUNA (red) relative to PPO (blue) and PPO with the prediction of $\FVE$ but without the filtering out of noisy samples (green) on 3 \textit{MuJoCo} environments averaged across 6 seeds. X-axis: number of environment steps. Y-axis: total undiscounted return. Shaded areas: standard deviation.}
  \label{fig:mujoco-nofilter-1}
\end{figure*}

We further study the impact of filtering out noisy samples by conducting additional experiments in predicting $\FVE$ while omitting the filtering step: the \textit{if} statement (\cref{line:if} of~\cref{alg:learning}) is removed and all transitions are kept in the batch ${\cal B}$. Indeed, SAUNA may improve the agent's performance by simply training the shared network to optimize the $\FVE$ head as an auxiliary task.~\cref{fig:mujoco-nofilter-1} demonstrates the positive effects of filtering out the samples. In addition, we studied the number of filtered out samples per task and its evolution along the training. On average, SAUNA rejects 5-10\% of samples at the beginning of training, 2-6\% near the end.

\subsubsection{The Impact of SAUNA on the Gradients}
\label{sec:grad}
Prior to the gradient update, SAUNA removes the useless transitions. By so doing, we hypothesized that information signals from samples with large $\FVE$ would be less diluted by filtering out samples.
\begin{figure}[h]
  \centering
  \subfloat[]{\includegraphics[width=.5\linewidth]{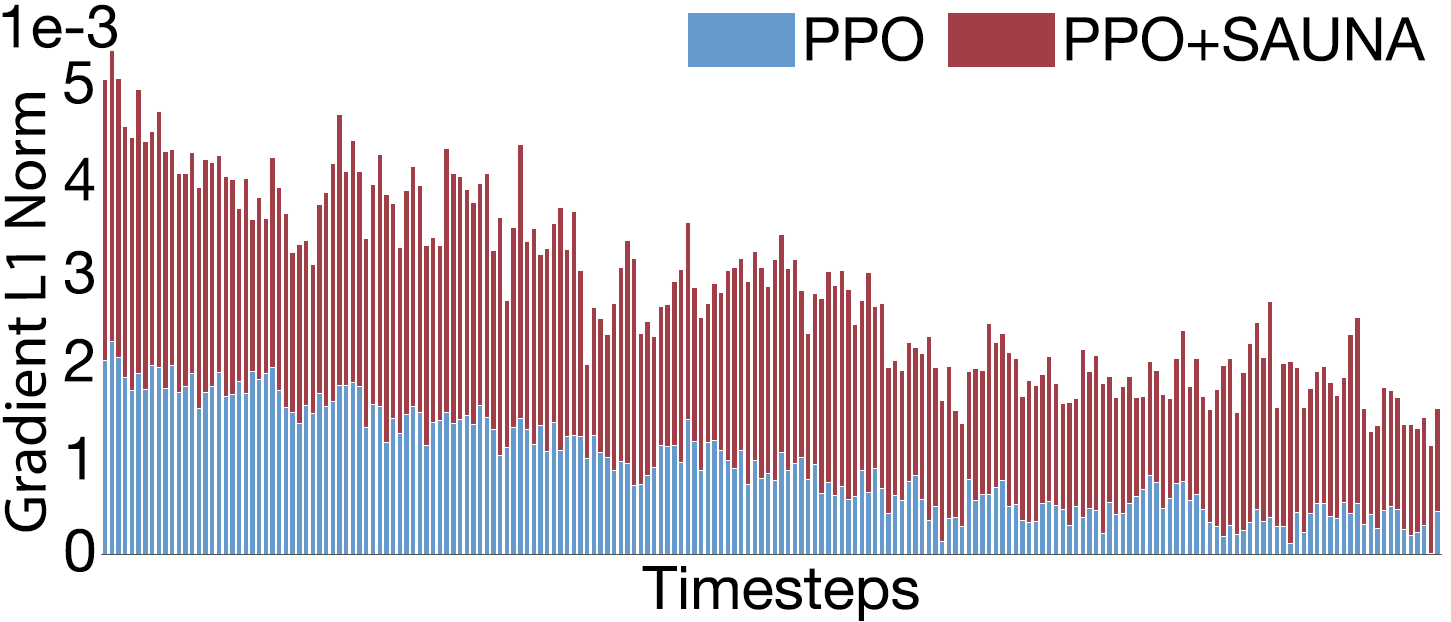}\label{fig:grad-first}}\hspace{-0.8cm}
  \qquad
  \subfloat[]{\includegraphics[width=.5\linewidth]{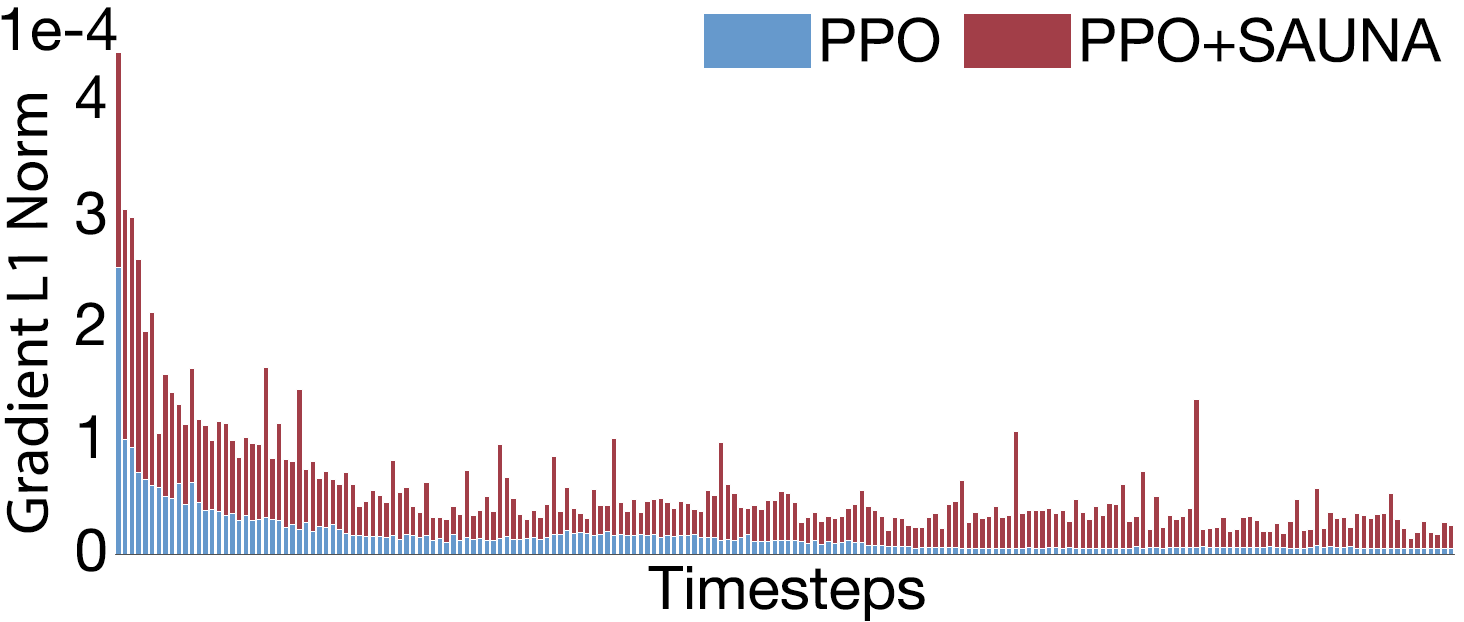}\label{fig:grad-last}}\\[-1ex]
  \caption{Gradients L1-norm from the (a) first layer and (b) last layer of the shared parameters network for PPO and PPO coupled with SAUNA. Task: \textit{HalfCheetah-v2}.}
  \label{fig:grad}
\end{figure}Fig.\@~\ref{fig:grad} shows that SAUNA filtering leads to larger gradients. As a result, policy updates make bigger steps, which ultimately translates into better performance. It is questionable why performance is not negatively affected, since larger gradients could hinder learning. Experience shows that gradients contain more useful information: as the relevant signals are less diluted, the gradients are more qualitative and have been partially denoised.

\subsubsection{\textit{HalfCheetah}: Qualitative Study}
\label{sec:case}
In \textit{HalfCheetah}, a well-known behavior~\cite{lapan_deep_2018} is that, for multiple seeds, a PPO agent gets stuck in a local minimum in which the agent moves on its back. However, we observed that SAUNA made it possible to leave from, or at least to avoid these local minima.
\begin{figure}[h]
  \centering
  \subfloat[]{\includegraphics[width=1.0\linewidth]{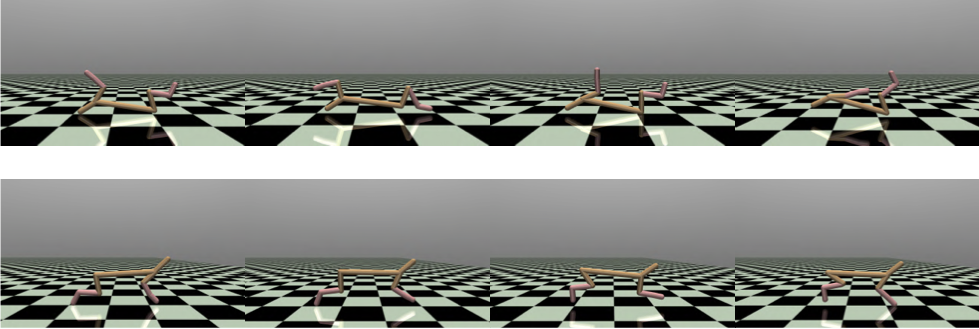}\label{fig:mujocoenv}}
  \qquad
  \subfloat[]{\includegraphics[width=1.0\linewidth]{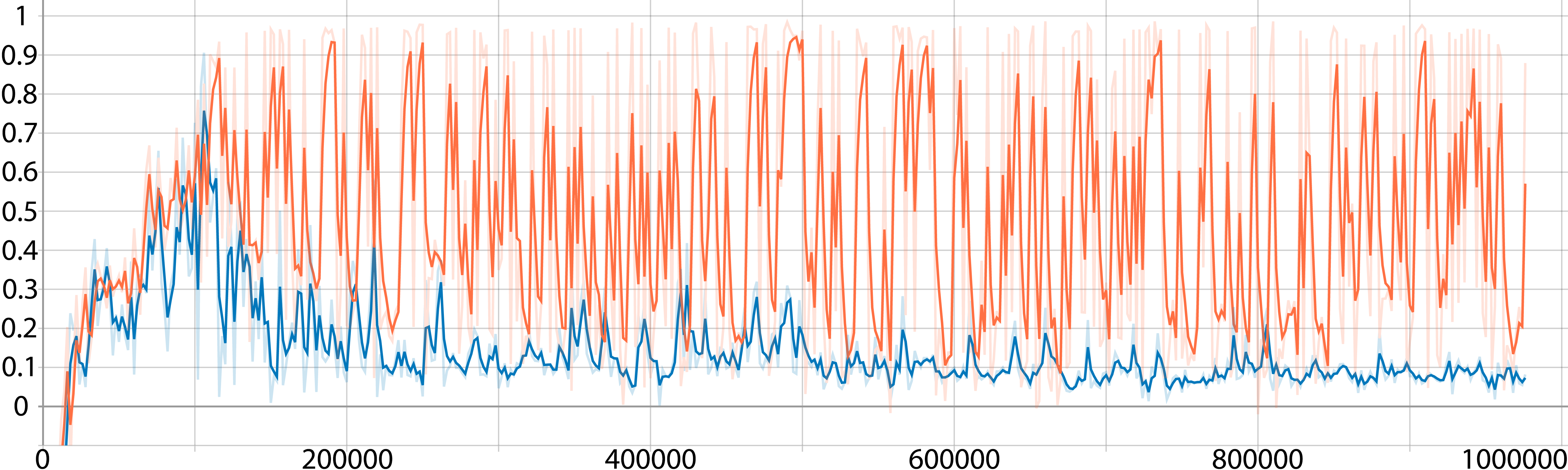}\label{fig:evresults}}\\[-1ex]
  \caption{(a) Example of PPO getting trapped in a local minimum (top row) while PPO+SAUNA reaches a better optimum (bottom row). (b) $\FVE$ score for PPO (blue) and PPO+SAUNA (orange).}
\end{figure}This is illustrated in~\cref{fig:mujocoenv} where we see still frames of two agents trained with PPO+SAUNA for $10^{6}$ timesteps on identically seeded environments. Their behavior is entirely different. Looking at $\FVE$ in~\cref{fig:evresults}, we can see that the graphs differ quite interestingly. The PPO agent seems to find very quickly a local minimum on its back while the blue agent's $\FVE$ varies much more. This seems to allow the latter to explore more states than the former and finally to find a better optimum. Supported by the previous study, we can infer that agents trained with SAUNA are better able to explore interesting states while exploiting with confidence the value given to the states observed so far.

\section{Discussion}
\label{sec:disc}
Intuitively, for the policy update, our method will only use qualitative samples that provide the agent with (a) reliable and exercised behavior (high $\FVE$) and (b) challenging states from the point of view of correctly predicting their value (low $\FVE$). SAUNA algorithm keeps samples with high learning impact, rejecting other noisy samples from the gradient update.

\subsection{Filtering Policy Gradient Updates and the Policy Gradient Theorem}
Policy gradient algorithms are backed by the policy gradient theorem~\cite{sutton2000policy}. As long as the asymptotic stationary regime is not reached, it is not reasonable to assume the sampled states to be independent and identically distributed (i.i.d.). Therefore, it seems intuitively better to ignore some of the samples for a certain period, to allow the most efficient use of information. One can understand SAUNA as making gradient updates more robust through filtering, especially when the update is low and the noise can be dominant. Besides, filtering out disturbing samples reduces the bias in the state distribution.

\subsection{Learning $\FVE$ and the Shared Network Parameters}
SAUNA network predicts $\FVE$ in conjunction with the value function and the policy. Therefore, as its parameters are updated through gradient ascent, they converge to one of the objective function minima (hopefully, a global minimum). This parameter configuration integrates $\FVE$, predicting how much the value function has fitted the observed samples, or informally speaking how well the value function is doing for state $s_{t}$. This new objective tends to lead the network to adjust predicting a quantity relevant for the task. Instead of using domain knowledge for the task, the method rather introduces problem knowledge by constraining the parameters directly.

\subsection{Additional Experimental Results}
We also compare SAUNA to A2C, a synchronous variant of~\cite{mnih2016asynchronous} and a weaker version of PPO. As expected, we observe a $15\%$ increase in performance. We do not present the complete results in this version of the article due to space limitations. Below is a discussion about additional experimental results, which we think contribute interestingly to the study.
\begin{description}[leftmargin=0pt]
  \item[Mean of $\FVE$.] Although $\widetilde{\FVE}$, the median of $\FVE$, is more expensive to calculate, we observe that it gives better results than if we use its mean in the \textit{if} statement of~\cref{alg:learning}. Using the median helps~\cite{10.2307/2683704} because the distribution of $\FVE$ is not normal and includes outliers that will potentially produce misleading results.
\end{description}
\begin{description}[leftmargin=0pt]
  \item[Non-empirical $\FVE$.] We also experimented with using the empirical values of $\FVE$ in \cref{line:if} of~\cref{alg:learning} when calculating $\widetilde{\FVE}$, instead of the predicted ones. This has yielded less positive results, and it is likely that this is due to the difference between the predicted and actual values at the beginning of learning, which has the effect of distorting the ratio in the \textit{if} statement.
\end{description}
\begin{description}[leftmargin=0pt]
  \item[Adjusting state count.] In order to stay in line with the policy gradient theorem~\cite{sutton2000policy}, we have worked to adjust the distribution of states $d^{\pi}$ to what it truly is, since some states visited by the agent are not included in the batch. We adjusted it using the ratio between the number of states visited and the actual number of transitions used in the gradient update, but this did not improve the learning, and instead, we observed a decrease in performance.
\end{description}
\begin{description}[leftmargin=0pt]
\item[Adjusted $\FVE$.] The definition of $\FVE$ is biased. An unbiased estimator does exist (known in statistics as the adjusted $R^2$). We performed the same set of experiments using such an adjusted $\FVE$: it did not change the experimental performance significantly.
\end{description}
\begin{description}[leftmargin=0pt]
  \item[Random filtering.] We experimented with dropping out at random, and before each gradient update, a number of samples corresponding to the same average number of samples that SAUNA drops. This resulted in a decrease in performance compared to PPO, as one can expect.
\end{description}
\begin{description}[leftmargin=0pt]
  \item[Atari domain.] We tested our method on the \textit{Atari 2600} domain~\cite{bellemare2013arcade} without observing any improvement in learning: some of the tasks were best performed by one method and others by the other.
\end{description}

\section{Conclusion}
\label{sec:conclusion}
Policy gradient methods optimize the policy directly through gradient ascent. We have introduced a new, lightweight and agnostic method applicable to any policy gradient algorithm. The central idea of this paper is that $\FVE$ is a useful measure to filter out samples that are perturbing the policy update.
Those non-informative or misinformative samples are ignored by SAUNA with a mechanism controlled by the estimated fraction of variance explained by the value function at each state. The relevant signals being less diluted, this improved sampling results in a denoising effect on the gradients, improving the learning curve, ultimately leading to improved performance.

We demonstrated the effectiveness of our method when applied to PPO, a commonly used state of the art policy gradient method, on a set of benchmark high-dimensional environments. We also established that samples can be removed from the gradient update without hindering learning but, on the opposite, can improve it. We further studied the positive impacts that such a modification in the sampling procedure has on learning. Several open topics warrant future study. Our results suggest that the influence of SAUNA on the distribution of states has beneficial effects: in order to gauge the theoretical implications of transition dropout in the MDP, our method might be formulated using the options framework~\cite{sutton1999between,precup2000temporal} where holes in a trajectory result in the appearance of options. Moreover, we are studying other ways to use $\FVE$ in the context of RL.

\bibliographystyle{named}
\bibliography{vex}

\end{document}